\documentclass{article}

\usepackage{acra}
\usepackage{graphicx}
\usepackage{multirow}
\usepackage[table,xcdraw]{xcolor}
\usepackage{adjustbox}
\usepackage{amsmath}
\usepackage[hyphens]{url}
\usepackage{subcaption}
\usepackage{hyperref}
\usepackage{siunitx}
\usepackage{wrapfig}
\usepackage{array}
\usepackage{multirow}
\usepackage{color, colortbl} 
\definecolor{Gray}{gray}{0.9} 

\title{Benchmarking Reinforcement Learning Methods\\ for Dexterous Robotic Manipulation with a Three-Fingered Gripper}
\author{
 Elizabeth Cutler$^{1}$, Yuning Xing$^{1}$, Tony Cui$^{1}$, Brendan Zhou$^{1}$, Koen van Rijnsoever$^{2}$, Ben Hart $^{2}$, \\\textbf{David Valencia$^{1}$, Lee Violet C. Ong$^{1}$, Trevor Gee$^{1}$, Minas Liarokapis$^{2}$, Henry Williams$^{1*}$}\\
    Centre for Automation and Robotic Engineering Science, The University of Auckland, New Zealand$^{1}$.\\
    New Dexterity research group, The University of Auckland, New Zealand$^{2}$.\\
  \texttt{henry.williams@auckland.ac.nz}$^*$ \\
}
\begin{document}

\maketitle

\begin{abstract}
    Reinforcement Learning (RL) training is predominantly conducted in cost-effective and controlled simulation environments. However, the transfer of these trained models to real-world tasks often presents unavoidable challenges. This research explores the direct training of RL algorithms in controlled yet realistic real-world settings for the execution of dexterous manipulation. The benchmarking results of three RL algorithms trained on intricate in-hand manipulation tasks within practical real-world contexts are presented. Our study not only demonstrates the practicality of RL training in authentic real-world scenarios, facilitating direct real-world applications, but also provides insights into the associated challenges and considerations. Additionally, our experiences with the employed experimental methods are shared, with the aim of empowering and engaging fellow researchers and practitioners in this dynamic field of robotics.

\end{abstract}

\section{Introduction}

    Dexterous robotic manipulation is the realm of robotics where multiple manipulators, typically robot fingers, collaborate to grip, handle, and manipulate objects within the hand \cite{844067}. In robotics, researchers typically use simple robot grippers and hands for the execution of complex industrial tasks like grabbing and moving components in structured environments \cite{tai2016state}. However, the varied tasks found in human-centred spaces like homes require adaptable manipulators like multi-fingered hands, which can execute a plethora of everyday actions such as moving objects, opening doors, and painting. Therefore, developing high-dimensional multi-fingered robotic grippers for dexterous manipulation has become an increasingly intriguing and unsolved challenge in robotics. 
    
    Controlling dexterous hands is a challenging task due to the required intricate coordination of sensor data, motor control, adaptability to object variability, and precise hand-object interactions \cite{ozawa2017grasp}. Implementing robot dexterity requires a combination of advanced hardware components, sophisticated software, and extensive development time to achieve success. Moreover, controlling a multiple-degree-of-freedom robotic hand with classical or heuristic controls can be impractical and complex due to the high level of control and fine-tuned motor skills required. Fortunately, Reinforcement Learning provides a more robust and adaptive solution for its ability to learn \cite{yu2022dexterous}. 
        
    RL holds the potential to automate complex control tasks by enabling agents to learn through trial and error. By defining a task and rewarding the agent accordingly, over time, the agent can learn to coordinate its joints to accomplish the task. RL training is commonly executed in simulation environments. However, challenges arise when transitioning these trained models into real-world applications.  Recently, there has been a growing research interest in conducting RL training in controlled real-world environments as shown in \cite{yu2022dexterous}. Other examples include \cite{ahn2020robel}\cite{valencia2023comparison} and \cite{zhu2019dexterous}. 
    
    The goal of this research paper is to add to the existing research into RL for robotic control and implementation of robotic dexterity by presenting: 
    \begin{enumerate}
        \item Benchmarking results on a real-world, three-fingered gripper to learn control behaviours.
        \item A general guide on conducting RL experiments in the real world for dexterous robotic manipulation. 
    \end{enumerate}
    
    \begin{figure}[ht]
        \centering
            \includegraphics[width=\linewidth]{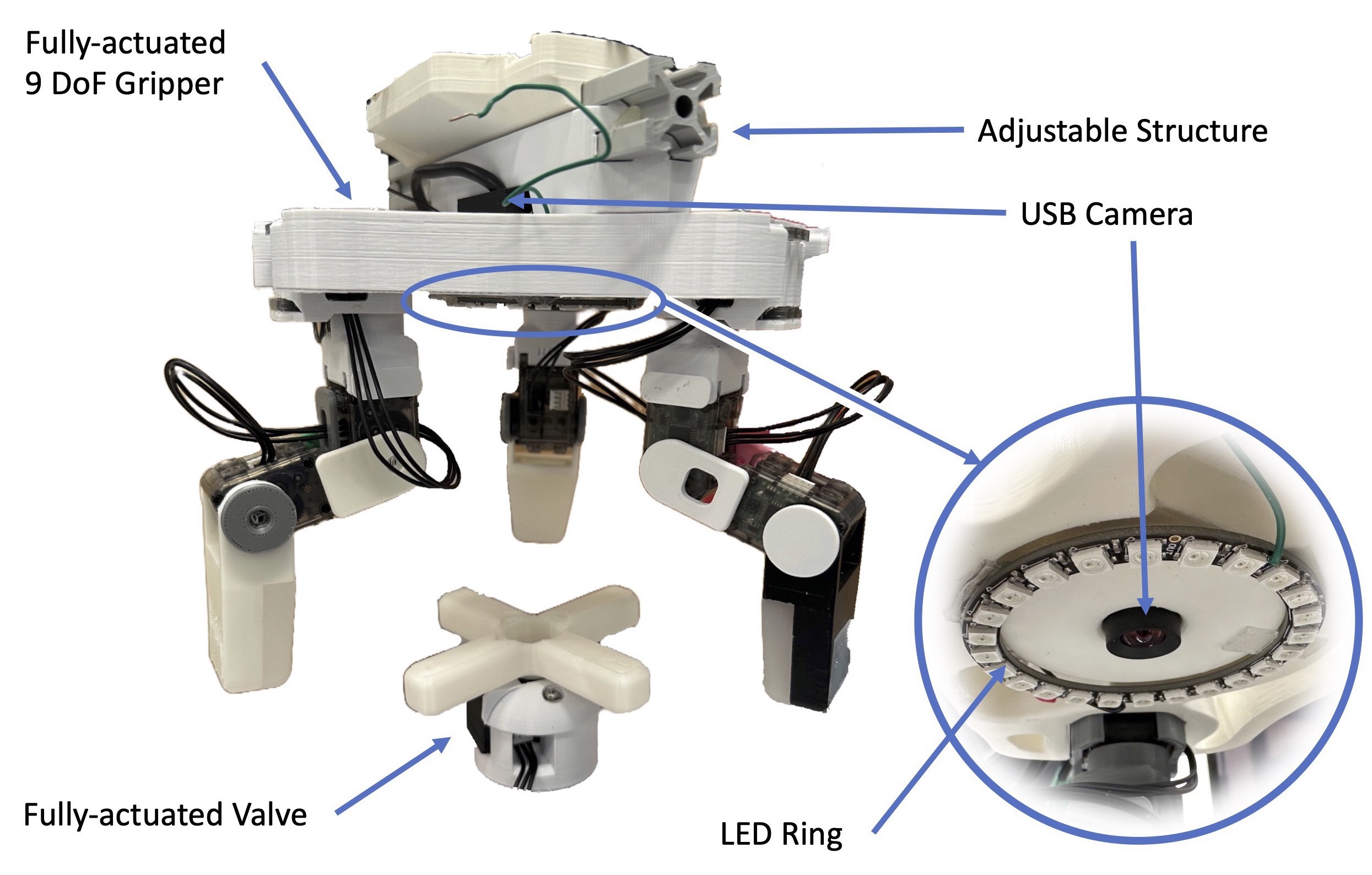}
            \caption{The employed fully-actuated 9 degrees of freedom robotic gripper with the rotational sensorized target valve. A USB camera is used to detect the valve's angle and for image-based RL training. An LED ring provides stable lighting. The gripper is 3D printed and attached to an adjustable structure.}
            \label{fig:gripper}
    \end{figure}

\section{Background and Related Work}
   
     To accurately present the reinforcement learning benchmark tasks that are assessed in this work on a real-world three-finger gripper, it is critical to understand the current state of the art of dexterous manipulator design, the RL training in the real world, and the current state of the art of RL for dexterous manipulation benchmarking.  

    \subsection{Gripper Design}
        Simulated manipulators are a popular training solution as there are no physical components to purchase or development time needed. Fatigue of actuators, cable management, and component failures are factors that do not need to be considered in the simulation. The ADROIT hand, used in \cite{fakoor2020ddpg} can complete the complex dexterous task of opening a door using all of the gripper's 24 degrees of freedom (DoF). However, simulated models are computationally expensive. For example, training the simulated Shadow Hand involved 384 worker machines, each with 16 CPU cores \cite{openai2019learning}. Even with such a large amount of computing power, the simulated robotic manipulators cannot be equipped with complete and accurate physics models. This leads to problems when transferring simulated control policies to the real world, known as the simulation-to-reality gap described in more detail in Section \ref{sec:sim-to-real}. 
        
        The design of dexterous robotic manipulators in the real world is also challenging, due to the several design considerations that need be taken into account: weight, cost, durability, accuracy, reliability, ease of use etc. The use of fully actuated designs with several motors is well known to increase the cost, weight, and complexity of the system, reducing the overall reliability. This has led to the development of underactuated dexterous robotic manipulators - where there are fewer actuators than the degrees of freedom - and the designs take advantage of structural compliance that allows them to conform to the object shapes and be more robust. In \cite{liarokapis2016learning}, the authors present an example of an underactuated end effector gripper that is used to learn new control actions in a supervised manner. Such an end-effector is easy to control but has limited fidelity, constraining the gripper to the execution of simple tasks. \cite{gupta2016learning} presents an extremely compliant gripper, more compliant than most, making the manipulator less likely to be damaged. The gripper is pneumatically actuated through 7 air chambers, is difficult to control with standard methods, and requires the use of a complex motion capture system so as to record the gripper kinematics and produce accurate results. Hence, the tasks that this manipulator can complete are limited by the corresponding control complexity. 
        
    \subsection{Real world RL training}\label{sec:sim-to-real}
        RL training is usually done in simulation as it provides a safe, cost-effective, easy to use, and efficient data collection solution. However, the challenges associated with bridging the simulation-to-real gap as presented in \cite{9308468}, pose a significant issue when applying trained models in real world environments and scenarios. This transition is difficult due to the assumptions used when creating a simulation that are unrealistic in real-world scenarios. These are phenomena that are difficult to model, such as friction between fingertips and the objects, uncontrolled contact slipping and rolling, and complex, non-uniform contact area compliance. Many attempts have been made to bridge this gap in the software realm. NVIDIA researchers incorporate domain, physics and non-physics randomisation into the simulation environment to better bridge this gap \cite{handa2023dextreme}. The challenges of real-world RL have been benchmarked and analysed in \cite{dulac2021challenges}. This paper presents nine challenges RL must succeed at before being deployed in real-world applications.
        
        To prevent this problem, training directly in the real world is being explored as an alternative. 
        \cite{ahn2020robel} introduced two robots designed to facilitate on-hardware RL training for dexterous manipulation (The 9 DoF robot D'Claw) and locomotion tasks (The 12 DoF robot D'Kitty). They also included a series of continuous control benchmark tasks tailored to each robot's capabilities.
        \cite{valencia2023comparison} compared model-free and model-based RL approaches directly in a real-world context. They accomplished this by training a low-cost four-DoF two-finger gripper. This experimental platform allowed the authors to evaluate the performance of RL algorithms in a real-world setting. 
        \cite{zhu2019dexterous} similarly demonstrated the feasibility of applying model-free RL algorithms to robotic hands. The proposed approach allowed them to successfully address various challenging tasks that were difficult to replicate in simulation environments. However, these are limited to low degree of freedom robotic grippers and the execution of simple tasks. This work increases the complexity of the gripper being employed and the tasks being solved. 

    \subsection{Dexterous Manipulator RL Benchmarks}
        
        The work in \cite{haarnoja2018soft} presents Soft-Actor Critic (SAC), an off-policy actor-critic algorithm aiming to simultaneously maximise expected return and entropy while acting as randomly as possible. It is evaluated by learning two challenging tasks directly in the real world, the first to teach a quadruped robot to walk on different terrains, and the second involves turning a valve using a 3-fingered robotic hand presented in \cite{zhu2019dexterous}. The results are impressive; however, the robot only rotates the valve to a fixed position. When only rotating the valve to a fixed position, the learned policy is essentially what the agent has memorised and considers as a strategy that will always work. If the initial position is changed slightly, the agent will have difficulty using its trained policy to complete the task. 

        The inclusion of offline data sets is studied in \cite{nair2020awac} and presents the advantage-weighted actor-critic (AWAC) method, enabling learning through prior demonstration data and online experience. Similarly, it is evaluated on a range of dexterous manipulation tasks in the real world, such as the three-fingered gripper rotating a valve precisely 180 degrees. It is shown to have a greater average return than all other prior work they studied, using offline and online data across 20k timesteps. However, the work is limited by setting a fixed value for rotation, and the gathering of prior knowledge is often difficult and impractical in the real world.

        A Model-Based Reinforcement Learning (MBRL) approach is presented in \cite{nagabandi2020deep} using online planning with deep dynamics models (PDDM). The task involves rotating two free-floating Baoding balls in the palm with a 24-DoF shadow hand. The rotation requires significant dexterity because the robot must find precise and coordinated manoeuvres to avoid dropping the objects. They found that PDDM outperforms prior model-based and model-free methods, which failed to succeed in learning. However, the authors use Model-Predictive control (MPC) \cite{lin2020comparison} to select actions instead of deep RL.
        
    Currently, the design of grippers and dexterous manipulators faces strong limitations. Many robotic grippers are under-actuated and compliant, lacking the fidelity required to complete complex enough dexterous tasks. Fully actuated real-world grippers can provide the dexterity required for the execution of complex tasks but often have to be simulated first to get good results coming at a greater cost than the often expensive gripper itself. Many of the existing grippers are also hard to replicate or purchase, with little to no open source material. Similar benchmarks are limited to rotating a valve to a fixed position, which may not demonstrate the full possibilities of the employed algorithm or gripper. 
    
    In this paper, we present the results of benchmarking tasks focusing on dexterous robotic manipulation, which are more complex than those previously presented, with several RL algorithms used with a three fingered gripper in the real world without the need for simulation - removing the sim-to-real problem. 
    
\section{Gripper}\label{sec:methods}

    This section provides a brief overview of the benchmarking testbed, presenting the employed gripper and the sensorized object.
    
    \subsection{Hardware Design}
        \subsubsection{Physical Design}
    
        Each robot finger is equipped with three degrees of freedom, two for flexion extension and one for abduction / adduction. Dynamixel XL-320 servo motors have been chosen as the base element for developing and motorizing the robotic fingers. This is because they meet the goal of making the fingers as modular and affordable as possible. For communication and power, XL-320s are able to be daisy-chained together, reducing the cable management complexity. These actuators make use of the Dynamixel SDK package for communication \footnote{\url{https://emanual.robotis.com/docs/en/software/dynamixel/dynamixel_sdk/overview/}\label{ft:dynamixelSDK}}, which makes it easy to access various registers of information within the servo. 
        The design was inspired by the ROBEL D’Claw \cite{ahn2020robel} as its design seemed suitable for the execution of complex dexterous manipulation tasks. The plane of rotation of the base joint is perpendicular to the planes of rotation of the two upper joints. The base of the fingertip is designed so as to mimic the rounded semi-spherical shape of a human fingertip and is aligned with the base joint. All three fingers are identical and almost entirely symmetrical, so if a gripper with more fingers needs to be developed, the only part that has to be redesigned is the gripper base plate. The base plate of the gripper has several options for how it is connected to its associated rig, making it usable for the execution of many complex tasks and for various applications. 

        \begin{figure}[ht!]
            \centering
            \includegraphics[width=\linewidth]{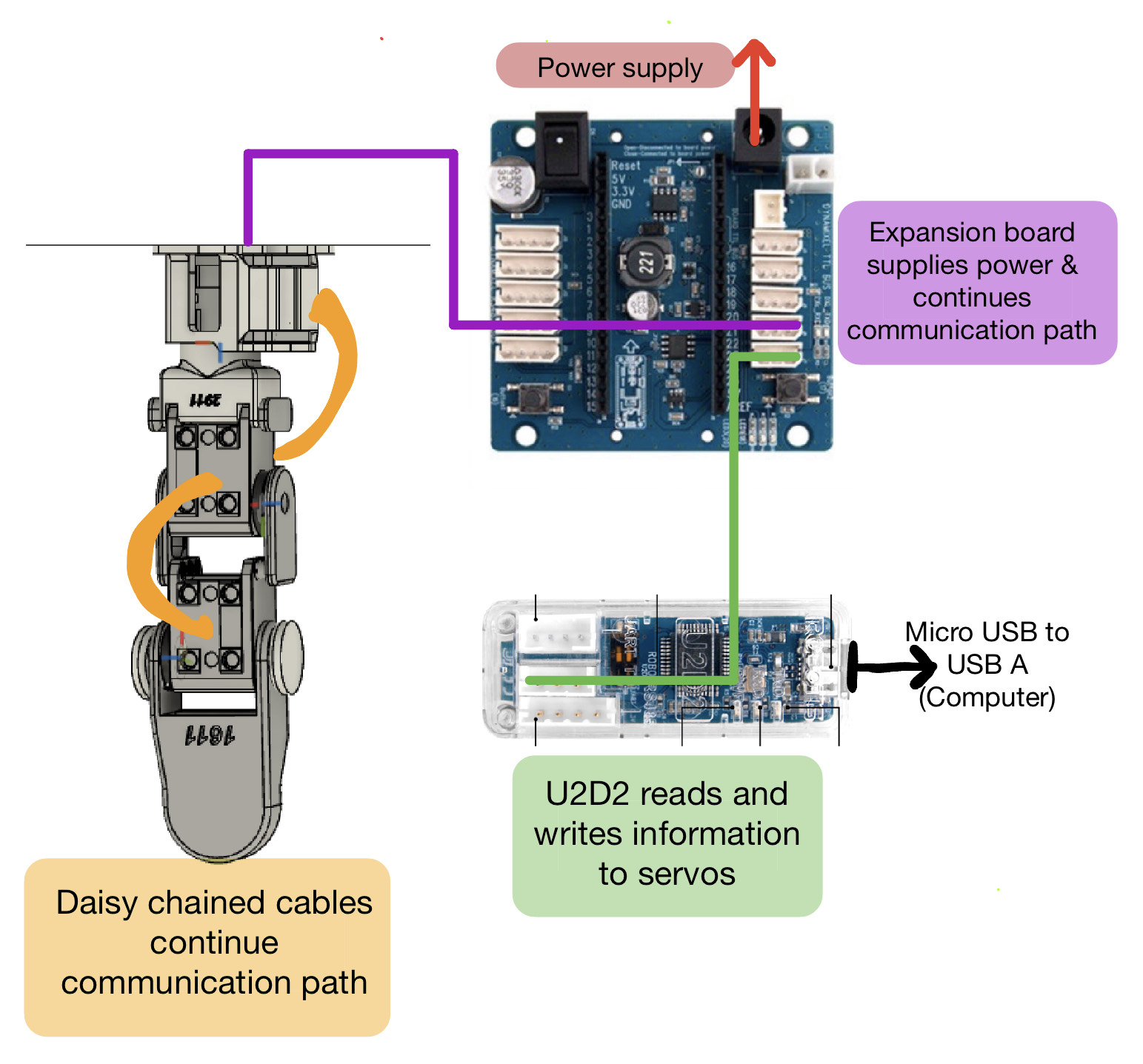}
            \caption{A summary of the electronic components and wiring for a single finger of the gripper}
            \label{fig:grippercircuitsetup}
        \end{figure}

        \subsubsection{Circuit Design}
        To communicate from the programme to the servos and from the servos to the programme, a Robotis U2D2 board is used. The U2D2 board provides both TTL and RS-485 communication protocols, but as the servos use TTL, only TTL is connected. The U2D2 board is connected to the Robotis OpenCM 485 Expansion Board, which supplies 7.7V of power and extends communication connections. Each finger is daisy chained independently, as the longer the daisy chain, the more susceptible it is to problems. 

        If the employed sensorized valve or another sensorized object requires communication with the programme and is run at 7.7V, it can simply be connected to the existing circuit. However, if it runs at a different voltage, a separate U2D2 and power source circuit is required. This experimental setup is summarised in Figure \ref{fig:grippercircuitsetup}.

    \subsection{Valve Observation Method}
        To complete the chosen benchmarking tasks, a sensorized valve is required. A 4-pronged valve similar to those found on taps and in \cite{ahn2020robel} was designed and 3d printed. 
        The angular position of the valve is required for the state vector, as described in section \ref{sec:stateactionspace}. To receive this position, the valve is connected to a Dynamixel XL330-M077-T servo motor that is equipped with a contactless absolute encoder (12-bit, 360°). This motor has a low stall torque, provides 0-360 degree angular position values, and uses the Dynamixel SDK package\textsuperscript{\ref{ft:dynamixelSDK}}, which allows it to integrate into the code easily.

\section{Experimental Methods} \label{sec:results}

    This work seeks to assess the efficiency of the model-free, deep RL algorithms in executing dexterous manipulation with a three-fingered gripper. To do so, the algorithms are trained and evaluated across three distinct rotational tasks. Each task is designed to mimic complex contact patterns and coordination demands reminiscent of everyday hand manipulations. To ensure the comparability of our results, we employ uniform sets of hyperparameters, state, and action spaces, and reward structures throughout our experimentation. This approach allows for fair and meaningful conclusions about the RL algorithms' adaptability and effectiveness across the execution of multiple tasks.

    \subsection{Tasks}\label{sec:tasks}
    A set of comprehensive evaluations have been conducted with the gripper, particularly focusing on its proficiency and adaptability in performing distinct tasks. These tasks are important for assessing the functional capacity of the gripper but also for determining its potential applications in real-world scenarios, especially where precision and adaptability are required. Each manipulation task assesses the algorithm's performance, encompassing training time, limitations, and real-world success rate. Three rotation tasks were defined, including a 90-degree rotation, a sequence of 90, 180, and 270 degrees, and rotations of 30 to 330 degrees. These tasks were designed to assess the algorithm's capacity to learn and generalise the objective of rotating the sensorized valve.

    The complexity of these tasks arises from the necessity for coordinated finger movements. To push and rotate the valve, the fingers must achieve a precise balance of torque and pressure. Adding to the challenge, the valve starts at a different angle between 0 and 360 degrees for each episode. This variability makes the training difficult because it prevents the network from relying on a memorised sequence of movements and compels genuine adaptation to each new task configuration.
    
    Dexterous robotic manipulation in real-world scenarios further intensifies these challenges. Unlike controlled environments, the real world presents unpredictable variables – variations in hardware functionality, external forces, and subtle shifts in positioning or alignment. A slight change in these environments can lead to entire task failures. Thus, we have generalised these tasks, extending and surpassing previous benchmarking tasks.

        \subsubsection{90 degrees}
        This task aims to turn the valve to an angle 90 degrees greater than the current angle. We employ the modulo operator to restrict the generated goal angle between 0 and 360 degrees to ensure a continuous range of goals. The successful completion of this task requires cooperative finger movements to push and rotate the valve, presenting an exploration challenge for the algorithms.
        
        \subsubsection{90, 180, 270 degrees}
        Like the 90-degree task, this task involves turning the valve to a goal angle of either 90, 180, or 270 degrees greater than the current angle. The additional complexity arises from the requirement to adapt to different destination angles rather than relying on a fixed set of movements that result in the same relative rotation.
        
        \subsubsection{30 - 330 degrees}
        The 30-330 degrees task represents the most challenging scenario. Here, the goal angle is randomly generated between 30 and 330 degrees relative to the current angle. Accomplishing this task successfully necessitates the acquisition of significant dexterity by the robotic gripper throughout the training process.     

    \subsection{Configurations}
        RL has made significant strides in enabling agents to learn from interactions within their environments. At the forefront of this progress are algorithms such as TD3 (Twin Delayed Deep Deterministic Policy Gradient) \protect\cite{fujimoto2018addressing}, DDPG (Deep Deterministic Policy Gradient) \protect\cite{lillicrap2015continuous}, and SAC \protect\cite{haarnoja2018soft}. These algorithms are specifically designed to handle challenges in continuous action spaces, making them relevant for robotic applications. 

        \begin{table}[]
            \centering
            \caption{Hyperparameters for all algorithms.}
            \label{tab:configs}
            \begin{tabular}{|l|l|}
                \hline
                Hyperparameters      &                            \\ \hline
                Actor Learning Rate  & $1 \times 10^{-4}$ \\
                Critic Learning Rate & $1 \times 10^{-3}$ \\
                Batch Size           & 32                         \\
                Buffer Capacity      & $1 \times 10^{6}$  \\
                G                    & 7                          \\
                Seed                 & 10                         \\
                Steps Per Episode    & 50                         \\
                Exploration Steps    & $1 \times 10^{3}$  \\
                Training Steps       & $6 \times 10^{4}$  \\ \hline
            \end{tabular}
        \end{table}

        For all the algorithms, the hyperparameters in Table \ref{tab:configs} were used to ensure consistent and unbiased training. Most of the parameters used are the default RL configurations for TD3, SAC, and DDPG. From testing and evaluation, the results indicate that a G value in the range of 7-8 yields the best training performance across all three algorithms. With a G value in this range, the algorithms demonstrated improved convergence rates and more stable learning trajectories.
        
        Exploration is a critical aspect that allows the model to try different actions to discover which ones yield the most rewarding outcomes. A value of 1,000 exploration steps was deliberately set to give the model sufficient opportunities to explore the action space without being overly biased by previous experiences and is flexible enough to converge to a sub-optimal strategy. 

        \begin{figure}[ht]
            \centering
            \includegraphics[width=\linewidth]{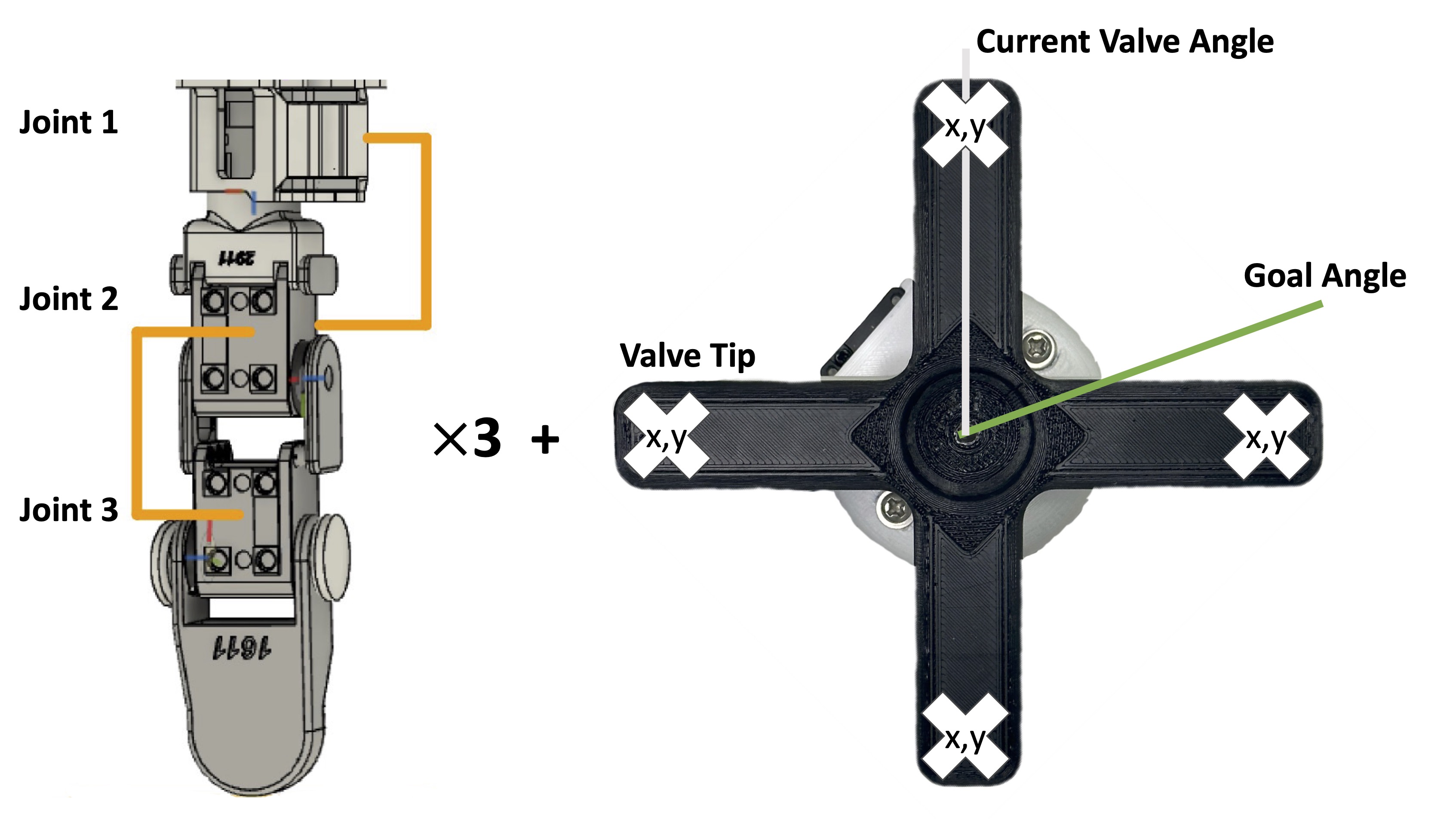}
            \caption{State space explanation.}
            \label{fig:states_illustration}
        \end{figure}

        When determining the optimal number of steps for training, both computational logic and hardware constraints play pivotal roles. In our case, we arrived at a strategic decision to set the limit at 60,000 steps during training. The rationale behind this wasn’t merely based on the theoretical or algorithmic side of things but also stemmed from practical constraints posed by the hardware. Given the structure of our training process, which incorporates evaluation steps, every training step not only requires forward and backward propagation for learning but also necessitates periodic evaluations to check the progress of the model.

        Evaluation steps, though essential for monitoring and ensuring the robustness of the training process, add an overhead to the computational time and resources. With each evaluation, the model's predictions are compared against a validation set, metrics are computed, and logs are stored for later analysis. As these frequent evaluations significantly extend the total training time.       
        Capping the training steps at 60,000, is an attempt to ensure that the model gets a sufficient amount of learning exposure while also leaving room for these periodic evaluations. Additionally, the set limit ensures that our training process remains not only computationally feasible, minimising the risk of potential resource overflows or long, unmanageable training times, but also time-efficient. This balance enables iteration, the ability to refine, and potentially retrain our model within a reasonable time frame, making experimentation more agile and responsive to any modifications or improvements.    

    \subsection{State and Action Space}\label{sec:stateactionspace}
        The observation state-space representation for the three-fingered gripper completing valve rotation tasks is a 1D vector consisting of 19 values. These values are the angle of the nine servo joints, the four-valve tips' positions (X and Y), the current valve angle, and the goal angle, as shown in Figure \ref{fig:states_illustration}. The four tip X-Y positions of the valve are calculated using trigonometry; these positions are provided for extra information on where the valve can be pushed. 
    
        A continuous action space is used to output a nine-element vector with the desired angle of each servo joint. The angle values produced are constrained between the maximum and minimum boundaries to prevent the servos from moving to physically impossible positions. 

        \begin{figure}[ht!]
            \centering
            \includegraphics[width=0.76\linewidth]{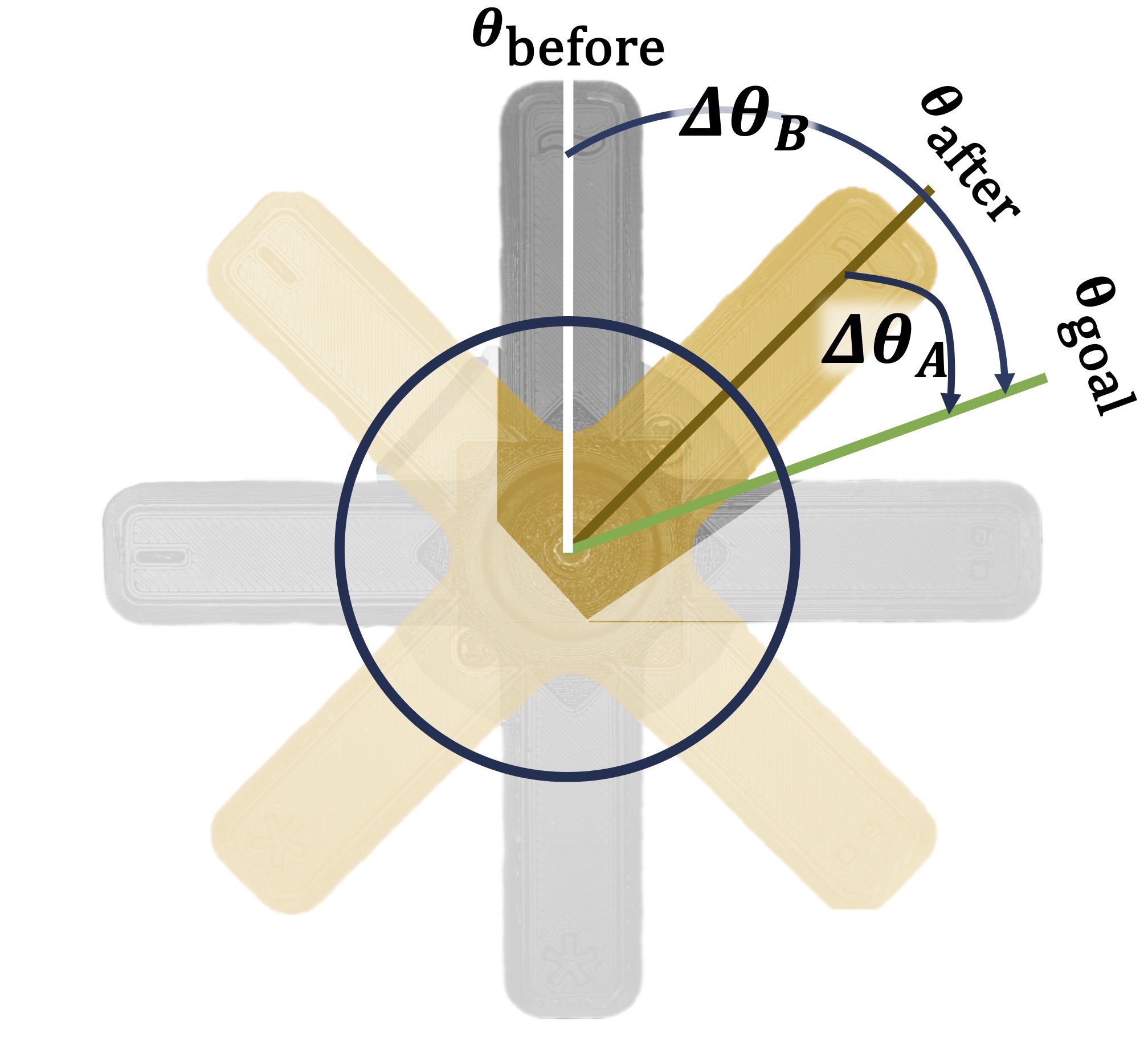}
            \caption{Reward function calculation illustration.}
            \label{fig:reward_function}
        \end{figure}

    \subsection{Reward Function}     
        A reward is given to the agent at each step, calculated using the following equations. 

        \[
            \Delta\theta = \Delta\theta_{B} - \Delta\theta_{A}
        \]
        \[
            R = 
            \begin{cases}
                 -1 , & \text{if } |\Delta\theta| < \epsilon\\
                 +10, & \text{elif  } |\Delta\theta_{A}|<\epsilon\\
                \Delta\theta / \Delta\theta_{B}, & \text{otherwise}
            \end{cases}
        \]
        
        As shown in Figure \ref{fig:reward_function}, $\Delta\theta_{B}$ and $\Delta\theta_{A}$ are the absolute difference between $\theta_{goal}$ and $\theta_{before}$/$\theta_{after}$. $\epsilon$ denotes the noise tolerance threshold, where any change or difference below this limit is considered negligible. If the goal has been reached before the episode concludes, an additional reward of 10 is awarded. 

        This approach to calculating rewards places emphasis on the agent's progress towards the goal, rather than penalising it for being far away. We argue this encourages the agent to focus on making progress and approaching the goal in a positive way. This approach can help prevent the agent from getting stuck or discouraged if it starts in a distant state, as it will receive rewards for any steps it takes in the right direction.
        
\section{Results}
\begin{figure*}[ht!]
        \centering
            \includegraphics[width=\textwidth]{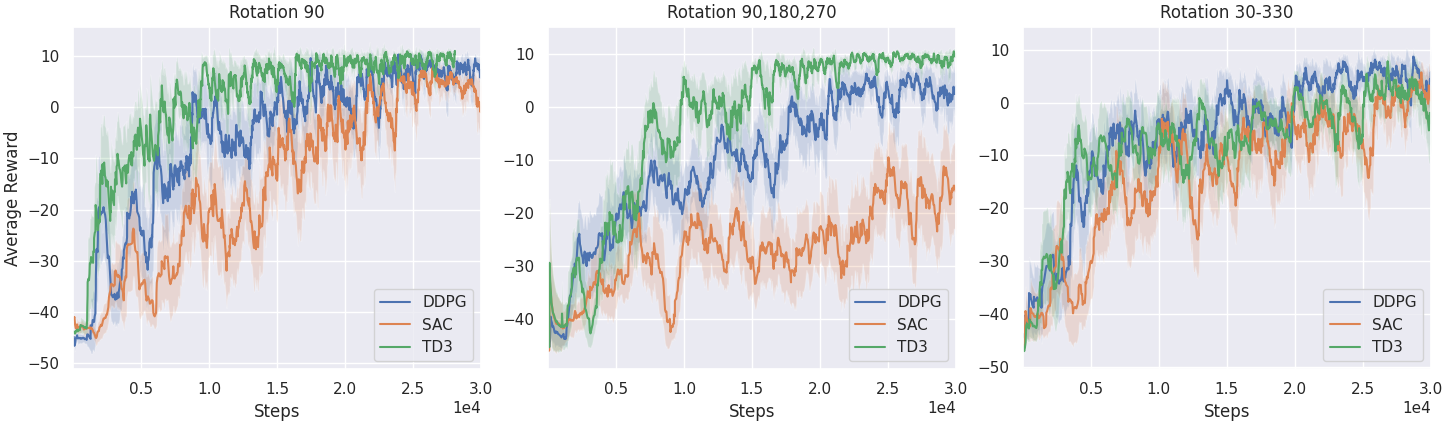}
            \caption{The learning curves of three model-free RL algorithms, Deep Deterministic Policy Gradient (DDPG), Soft Actor Critic (SAC), and Twin Delayed DDPG (TD3) were evaluated on the three tasks. Reward at each step is determined by how much closer the gripper moves the valve towards the goal angle.}
            \label{fig:reward_plot}
    \end{figure*}

    Figure \ref{fig:reward_plot} presents a comparative evaluation of the learning processes employed by three model-free RL algorithms: DDPG, SAC, and TD3. The primary metric for comparison is the average reward evaluation. Figure \ref{fig:reward_plot} further delineates the performance of each algorithm on each executed task. 

    \begin{figure*}[h!]
        \centering
            \includegraphics[width=\linewidth]{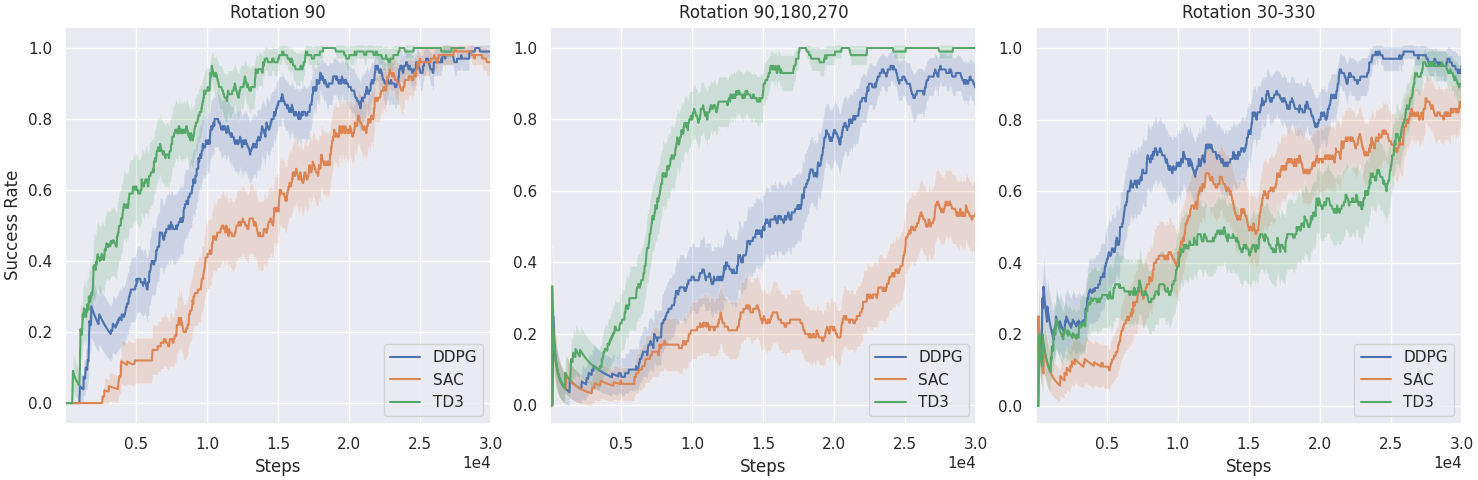}
            \caption{Success rate for the three tasks trained on multiple physical grippers using the three algorithms. A success rate of 1 means that the gripper reached the goal angle within one episode (50 steps).}
            \label{fig:success plot}
    \end{figure*}

    In Figure \ref{fig:success plot}, we present the success rates of three tasks, each trained on multiple iterations of dexterous manipulation executed with the three-fingered gripper using three distinguished algorithms: DDPG, SAC, and TD3. The success rate metric serves as a vital tool to gauge the effectiveness and efficiency of each algorithm when applied to physical hardware. Importantly, a success rate of 1 indicates that the algorithm, within a single episode of 50 steps, successfully guided the gripper to its intended goal angle without error. This not only highlights the algorithm's precision but also its responsiveness, especially when integrated with cost-effective, real-world hardware. Comparing these success rates across various tasks and algorithms offers a deeper insight into the advantages and potential challenges each algorithm presents in practical settings.
    
\begin{figure*}[h!]
        \centering
            \includegraphics[width=\linewidth]{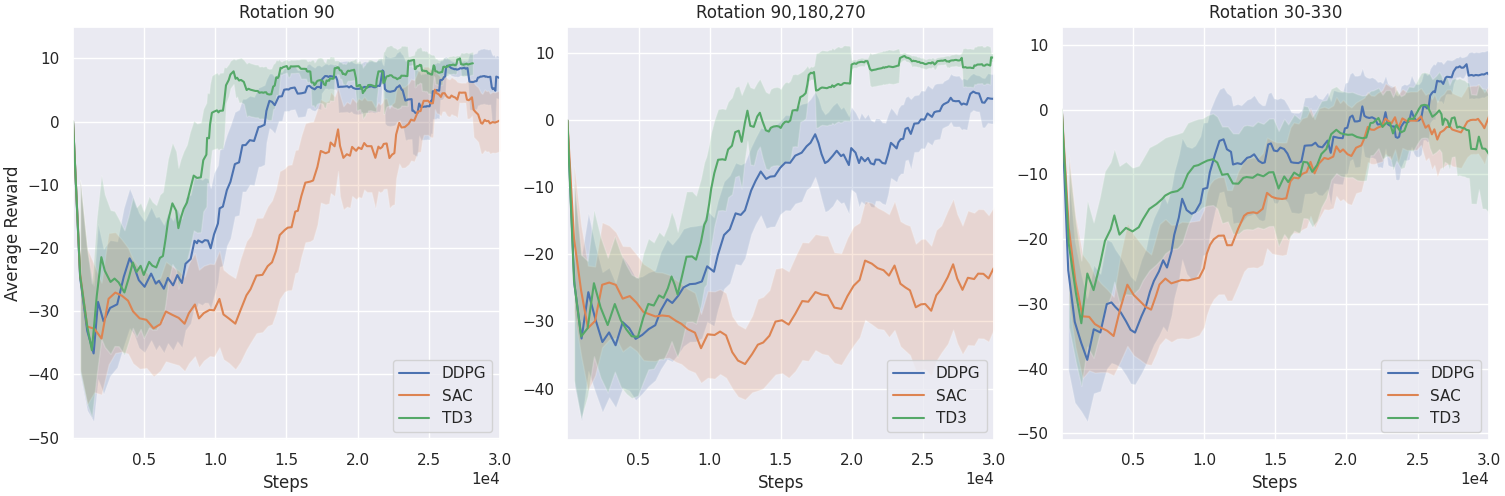}
            \caption{The agent was evaluated by running 1 evaluation episode (50 steps) every 10 training episodes. In evaluation mode, the agent does not update the policy. The average reward over the evaluation episode was computed and plotted.}
            \label{fig:evaluation plot}
    \end{figure*}

    Figure \ref{fig:evaluation plot} showcases the evaluation process of the agent. The agent was evaluated through an evaluation episode lasting 50 steps, every 10 training episodes. Notably, during these evaluation phases, the agent remained static in its approach, refraining from learning or improving its policy. The results represent the computed average reward over each evaluation episode. From the displayed outcomes, it is evident that TD3 consistently outperforms its counterparts. This observation highlights the robustness and efficiency of the TD3 algorithm in the efficient execution of the considered tasks in the relevant environments. 
    To demonstrate the gripper's capability in task execution, an evaluation loop is conducted using the final model of each trained agent, encompassing 1000 steps. Performance assessment was based on success metrics as previously defined, with results shown in Table \ref{table:eval_after_training}. Across all three tasks, a recurring pattern emerged, with TD3 consistently achieving a success rate exceeding 90\%. DDPG and SAC closely trailed behind TD3 in performance.
    
     The reason behind TD3's better performance is likely multifaceted. At its core, TD3 employs a twin-delayed deep deterministic policy gradient approach. This methodology aims to tackle the challenges of value overestimation. One key strategy TD3 employs to reduce overestimation is by using two critics and taking the minimum value of the two, which effectively addresses the issue of over-optimistic value estimates. This is in addition to updating the policy less frequently than the Q-values and using noise regularisation. Furthermore, TD3 incorporates a policy smoothing technique to combat the extrapolation error by adding noise to the target policy during the critic update. These architectural and algorithmic improvements, combined with a more stable learning process, potentially render TD3 as the standout among the trio in the execution of the dexterous robotic manipulation tasks and contexts showcased in figures 6, 7 and 8.\\

    A video demonstrating the training process and performance of the final policy for different tasks can be found \href{https://youtu.be/0kii1EJjOzw}{here}\footnote{\url{https://youtu.be/0kii1EJjOzw}}.
    
    \begin{table}[]
        \centering
        \begin{tabular}{|c|ccc|}
            \hline
            \multirow{2}{*}{Algorithm} & \multicolumn{3}{c|}{Task}  \\ \cline{2-4} 
                                       & 90   & 90,180,270 & 30-330 \\ \hline
            TD3                        & 1.00 & 0.99       & 0.93   \\
            DDPG                       & 0.87 & 0.93       & 0.91   \\
            SAC                        & 0.88 & 0.65       & 0.55   \\ \hline
        \end{tabular}
        \caption{Success rate of the final trained RL models in successfully completing the execution of the dexterous robotic manipulation tasks.}
        \label{table:eval_after_training}
    \end{table}

\section{Discussion}
    The evaluation of the DDPG, SAC, and TD3 methods, as illustrated through various metrics and visual representations, offers valuable insights into the relative learnings of these model-free RL algorithms when piloted in the execution of complex dexterous manipulation tasks with physical hardware like the employed three-fingered robotic gripper.

    TD3 is the best and emerges as the leader in terms of performance. This is particularly evident when considering the learning curves, wherein TD3 consistently showcases a more stable and rapid convergence towards optimal solutions. Such a performance underscores its innate robustness and capability to efficiently navigate the complexities associated with the execution of complex, dexterous real-world tasks. This efficiency may well be attributed to the twin-delayed approach and judicious use of two critics, minimising pitfalls such as value overestimation, a notorious challenge in RL. 
    
    Although TD3 proves to be the best, it's pivotal not to overshadow the potential of both the DDPG and the SAC RL methods. DDPG, despite its occasional variances, demonstrates a laudable ability to drive physical robot hardware towards desired outcomes. On the other hand, SAC's resilience, albeit sometimes converging more gradually, underlines its versatility in fluctuating conditions.

    The success rates, while offering a direct reflection of precision, also elucidate the algorithms' adaptability in dynamic, real-world settings. TD3's frequent inclinations towards optimal success rates accentuate its predominant position. However, the comparable achievements of DDPG and SAC indicate their applicability and potential superiority, contingent on specific tasks or environments. Whilst TD3 is often favoured in the evaluated benchmarks, the distinct contributions and potential of DDPG and SAC must be evaluated further. The findings presented serve not just as a testament to the algorithms' capabilities but also as a guide for the application of the three-fingered robotic gripper in more diverse real-world scenarios.
    
    When implementing training in real-world scenarios as opposed to controlled, simulated environments, several pressing concerns arise. One of the most significant is the limitations associated with gripper hardware. Over time and through the repeated use inherent to RL training, the components of grippers exhibit inconsistencies due to wear and tear. Continuous motion and external intrusions like dust or debris lead to deviations in expected behaviour. The communication from the software to the hardware and back can also break down, with issues such as status packets that are not recognised or servos not being able to be identified. When such incidents occur, not only is the training session compromised, but the data accumulated might also be rendered unusable. Training sessions may need to be restarted from the beginning to ensure data integrity. This reset, if unable to be fixed quickly, undoubtedly affects overall efficiency. 

    Navigating the challenges of the project required innovative solutions, particularly when addressing the recurring issues associated with the gripper's operational stability. A primary concern was the frequent necessity to reboot or restart the gripper. Originally, this process was executed manually, resulting in potential disruptions and extended operational downtime. To alleviate this, an automated solution was introduced: a function designed to automatically reboot the gripper upon failure. This function would attempt multiple reboots. After these systematic attempts, if the gripper failed to respond appropriately, the system was programmed to revert to a manual reboot mode. This strategic implementation of a hybrid approach aimed to streamline operations, minimise manual interventions and enhance the overall efficiency of the project's testing and implementation phases.   
    
    Initial prototypes of the setup used ArUco markers paired with cameras to gauge the valve's rotational angle. This method proved challenging as the cameras' sensitivity to lighting conditions led to inconsistencies in angle detection. Particularly during low-light scenarios, the cameras struggled to discern the ArUco markers accurately, causing discrepancies in the recorded measurements. The setup then transitioned to using a magnet for resetting the magnetized object combined with an encoder. While this setup provided a more consistent measurement compared to the previous method, it was not without its challenges. The primary difficulty comes from the extended data retrieval times, as the inherent delay in reading from the encoder often resulted in a lag when attempting to retrieve the valve's angle in real-time. This lag became problematic especially in scenarios requiring rapid feedback and timely responses, hampering the overall effectiveness of the system and causing some training data to be unusable.
    
    To resolve the issues presented in the magnetic encoder, a Dynamixel servo was adopted for monitoring the valve's angle. This transition marked a significant improvement in the setup's efficiency. The Dynamixel servo with its built-in contactless absolute encoder, increased the precision of angle readings in real-time, eliminating the delays previously encountered. Additionally, the servo's inherent reliability ensured a high degree of consistency in measurements, streamlining the process and significantly enhancing the project's overall precision and reliability.

\section{Conclusions and Future Work}
    Benchmarking results indicate that TD3 consistently outperforms both DDPG and SAC across all tasks, underscoring its efficacy and resilience in addressing real-world continuous tasks. Throughout the series of benchmarking tasks presented, RL training in the real world has revealed its practicality and advantages, effectively mitigating the simulation-to-real gap. Despite grappling with hardware issues related to wear and tear, the trained model has demonstrated robustness and adaptability to diverse conditions. This paves the way for the potential of training RL models in executing complex tasks in controlled real-world environments.

    Future work will focus on implementing velocity-based control methods into the gripper, in contrast to the current method of position control. As part of this, we will introduce tactile sensors into the fingertips of the three-fingered gripper that will be utilised to detect when the gripper touches the valve. With this addition, the expectation is to observe an improvement in the training of the three-fingered gripper as the sensors will provide an additional set of information that can be used as a reward. Furthermore, velocity-based control is expected to allow for more dynamic motion of the gripper, which may potentially give way to higher success in executing complex, dexterous  robotic manipulation tasks.

\section*{Acknowledgements}
    This research was supported by New Zealand's Science For Technical Innovation (SfTI) on contract UoA3727019.

\bibliography{publications}
\bibliographystyle{named}
\end{document}